# LLM Based Sentiment Classification From Bangladesh E-Commerce Reviews

Sumaiya Tabassum
*Department of Computer Science and Engineering*
*Dhaka International University*
Dhaka, Bangladesh
sumaiyatabassum230@gmail.com

*Abstract*—Sentiment analysis is an essential part of text analysis, which is a larger field that includes determining and evaluating the author's emotional state. This method is essential since it makes it easier to comprehend consumers' feelings, viewpoints, and preferences holistically. The introduction of large language models (LLMs), such as Llama, has greatly increased the availability of cutting-edge model applications, such as sentiment analysis. However, accurate sentiment analysis is hampered by the intricacy of written language and the diversity of languages used in evaluations. The viability of using transformer-based BERT models and other LLMs for sentiment analysis from Bangladesh e-commerce reviews is investigated in this paper. A subset of 4000 samples from the original dataset of Bangla and English customer reviews was utilized to fine-tune the model. The fine-tuned Llama-3.1-8B model outperformed other fine-tuned models, including Phi-3.5-mini-instruct, Mistral-7B-v0.1, DistilBERT-multilingual, mBERT, and XLM-R-base, with an overall accuracy, precision, recall, and F1 score of 95.5%, 93%, 88%, 90%. The study emphasizes how parameter-efficient fine-tuning methods (LoRA and PEFT) can lower computational overhead and make it appropriate for contexts with limited resources. The results show how LLMs can help advance sentiment analysis for low-resource languages.

## I. INTRODUCTION

The explosive growth of e-commerce platforms has caused a major shift in global trade in recent decades. Product reviews have become a crucial component in determining purchasing decisions and forming brand perceptions as a result of customers' increasing reliance on internet shopping [1]. Customers may learn a lot from these reviews, which include information about user contentment, product quality, and overall satisfaction. Sentiment analysis has seen significant study attention as a result of the growing significance of comprehending the opinions stated in these reviews.

The technique of identifying the underlying feelings or viewpoints in textual data, such as whether the sentiment is neutral, negative, or positive, is known as sentiment analysis. However, there are many difficulties in determining the emotion of online product reviews[2]. One significant barrier is that people frequently disregard stringent grammar rules while writing online. Significant complication is introduced by the informal and unstructured character of online communication, which includes comments, product evaluations, and social media interactions. It might be challenging to precisely capture and comprehend the thoughts expressed by customers in these communications since they frequently reflect a heterogeneous linguistic landscape. One of the main goals of Natural Language Processing (NLP) research has been to identify and treat these linguistic variances. There is still a lack of knowledge on the linguistic categories that predominate in Bangladeshi product reviews, despite the fact that numerous qualitative studies have looked at the language dynamics of Bangla speakers in online communication[3]. Prior studies on sentiment analysis in Bangla, Banglish, and code-mixed Bangla-English texts have investigated a variety of deep learning and machine learning (ML) techniques, such as convolutional neural networks (CNN), long short-term memory (LSTM) networks, and support vector machines (SVM). But comparatively few of these studies examine Bangla-English code-mixed text, which combines both Bangla and English scripts, with the majority concentrating on romanized Bangla. This dearth of thorough research emphasizes the necessity of a more complete strategy to comprehend the linguistic environment in this setting and enhance sentiment analysis methods for reviews in mixed languages[4]. Significant strides have been made in the development of LLMs within the ever-evolving discipline of NLP, leading to a noteworthy paradigm change. LLMs use conceptual analysis to capture significant facets of meaning by conceptual roles, which can explain their accomplishments and offer suggestions for how to humanize them. Transformer language models with hundreds of billions or more of parameters are frequently referred to as LLMs. Furthermore, there is a great chance for additional study and advancement because the potential of LLMs in the Bangladeshi context is still mainly untapped. To address these challenges, this study introduces a sentiment analysis framework that combines automatic language categorization with LLMs and BERT models to enhance sentiment prediction into relevant linguistic categories—such as Bangla, Banglish, Bangla-English, and English. By integrating language classification techniques with LLMs, this research aims to explore the potential of Llama in bangla sentiment analysis, optimize its computational efficiency, and validate its performance against existing models. The study is capable of handling the diverse linguistic nature of Bangladeshi product reviews, offering a robust solution for accurate sentiment prediction. The primary contributions of the proposed research can be summarized as follows:

1) We investigated the linguistic landscape of Bangladeshi product reviews, exploring the prevalence and dominance of various language categories, including Bangla, English, Banglish, and Bangla-English code-mixed. This study addresses these challenges by leveraging Llama, an advanced LLM, to bridge the gap in NLP research of about Bangla and English mixed product reviews for sentiment analysis.

2) We analyzed the effectiveness of LLMs' for sentiment analysis in Bangla & English mixed language by evaluating the performance using five open-source LLMs & BERT models such as Llama-3.1-8B , Phi-3.5-mini-instruct, Mistral-7B-v0.1, DistilBERT-multilingual, mBERT, and XLM-R-base, across different language categories, providing a comprehensive comparison of their effectiveness.
3) We assessed the performance of base LLMs, BERT models in terms of accuracy, precision, recall, and F1 score, demonstrating the superior performance and versatility in handling mixed-language data.

The rest of this paper is structured as follows: Section II provides a detailed review of related literature. In Section III, we present the methodology, Section IV offers a comparative analysis of the results obtained from implementing various LLM & BERT models. Finally, Section V concludes the study with statements of the limitations of this study and discusses potential future directions for research.

## II. LITERATURE REVIEW

First, Different languages, code-mixing, and writing styles make multilingual sentiment analysis difficult, particularly in places like Bangladesh, where evaluations frequently blend Bangla, English, and transliterations. These multilingual and mixed-text reviews can now be handled more accurately and efficiently and thanks to recent advancements in LLMs.The majority of Bangladeshis speak Bangla as their mother tongue, and it is the country's official language. Numerous Bangla Natural Language Processing efforts (BNLP) have been carried out recently. Bangla sentiment analysis has greatly improved because of deep learning techniques like LSTM, GRU, and Bangla-BERT. Accuracy and emotional comprehension have increased thanks to refined transformer models. LLMs have recently sparked a paradigm change in NLP, demonstrating great promise in a variety of fields, including sentiment analysis. This project aims to improve sentiment classification in Bangla utilizing state-of-the-art LLMs by fine-tuning on a custom e-commerce dataset, as their application to Bangla sentiment is still limited.

Recent research emphasizes how difficult sentiment analysis is in multilingual settings like Bangladesh, where evaluations frequently contain code-switched language in addition to English, Bangla, and Banglish. To tackle this issue, BdSentiLLM was developed, combining sentiment analysis and language classification, and it was tested with Llama-2, Flan-T5, Vicuna, and Falcon. The results demonstrated that BdSentiLLM with Llama-2 outperformed current models in handling mixed-language evaluations, achieving the best overall performance (F1 = 0.89)[5]. Sentiment analysis is crucial for understanding user perspectives, but research on low-resource languages like Bangla is still lacking. A recent study applied Llama with parameter-efficient fine-tuning (LoRA, PEFT) to 1,000 Bangla reviews, effectively addressing resource restrictions. The improved model outperformed Bangla-BERT by 1% with an accuracy of 89.76%, indicating the promise of LLMs for sentiment tasks with lower resource requirements[6].Despite its significance in NLP, there is still little research in Bengali sentiment analysis in comparison to English. A recent study used resampling strategies to overcome class imbalance and improve BanglaBERT and mBERT on a bespoke Facebook comments dataset. The models outperformed previous standards in Bengali sentiment analysis, with BanglaBERT achieving 86% accuracy and mBERT 87%[7].According to recent research, sentiment analysis based on LLM can improve the classification of financial news for stock prediction. Llama-3 is the recommended model for sentiment estimation since it produced stable results in comparisons, outperforming FinBERT and VADER. Forecasting accuracy increased when these sentiments were taken into account, with Informer performing best in the majority of instances[8].Bengali political speech is difficult to categorize because of its linguistic complexity, lack of resources, and domain-specific variances. According to a study evaluating seven transformer-based models, RoBERTa outperformed BERT in terms of accuracy (91%), demonstrating their ability to capture subtleties in context. These findings support the usefulness of transformer designs, particularly in political text analysis, for low-resource languages like Bengali.[9]Past works on Bangla sentiment analysis primarily utilized transformed conventional machine learning models (such SVM and KNN) and models based on transformers, such as Bangla-BERT. Although these methods produced notable gains, they frequently needed a lot of processing power and had trouble with managing Bangla's complex grammatical structure. For example, Bangla-BERT showed excellent accuracy but was not scalable in environments with limited resources. This work uses Llama, which leverages parameter-efficient fine-tuning techniques to improve computational efficiency without sacrificing accuracy, in contrast to earlier approaches. This study is unique because it addresses the practical shortcomings of previous models by using the LoRA and PEFT approaches. Unlike previous methods, this study fills a research gap in Bangla, English mixed NLP by presenting an accurate, scalable, and efficient LLM-based solution for sentiment analysis.

## III. METHODOLOGY

### A. Dataset Description

The dataset comprises of 78,130 product reviews from 152 product categories on two of the biggest e-commerce sites in Bangladesh, Daraz and Pickaboo[10]. An extensive resource for sentiment and emotion analysis in the Bangladeshi market, the reviews are written in both Bengali and English and are categorized into two sentiment classes (Positive and Negative) and five emotions (Happiness, Sadness, Fear, Anger, and Love). The dataset includes columns for Rating, Review text, Product Name, Product Category, annotated Emotion, overall Sentiment, and Data Source. The Overview of sample data and column structure in the dataset CSV is given at TABLE I. Reviews were published in both Bengali and English.A subset of 4,000 samples from the original dataset was used in each experiment. Only binary sentiment classification (positive vs. negative) was the subject of the analysis. Sentiment labels were encoded as integers: Negative(0), Positive(1), and this study used the review and sentiment columns for classification of sentiment analysis. In an 80:10:10 ratio, the subset of the dataset was divided into training, validation, and test sets after being shuffled. To guarantee the reproducibility of the sampling and splitting, a fixed random seed was employed. A balanced distribution of sentiment classes across all splits was guaranteed by this stratified sample.

## B. Research Methodology

Two types of experiments were conducted: one that employed LLMs and the other that used BERT-based models. With an emphasis on their capacity to capture syntactic and contextual subtleties in a low-resource environment, the BERT-based models were specifically adjusted for the Bangla sentiment dataset. The LLM-based tests, on the other hand, assessed the performance of cutting-edge generative models. A comparative viewpoint on the efficacy of large-scale, general-purpose models and lightweight, domain-adapted models was made possible by this dual approach.

TABLE I. THE OVERVIEW OF SAMPLE DATA AND COLUMN STRUCTURE IN THE DATASET

|   | Rating | Review | Product Name | Product Category | Emotion | Data Source | Sentiment |
|---|---|---|---|---|---|---|---|
| 0 | 5.0 | অসাধারণ ফোন।অনেক পছন্দ হয়েছে… | Redmi 12C (4/128GB) | Smart Phones | Happy | Daraz | Positive |
| 1 | 5.0 | Phone is good according to my uses, Upgraded… | Redmi 12C (4/128GB) | Smart Phones | Happy | Daraz | Positive |
| 2 | 5.0 | অল্প দামে দারুন একটা স্মার্টফোন | Redmi 12C (4/128GB) | Smart Phones | Love | Daraz | Positive |
| 3 | 5.0 | Super Fast Delivery, 11200 TK te pailam | Redmi 12C (4/128GB) | Smart Phones | Happy | Daraz | Positive |

### 1) Large Language Models

The process for optimizing and assessing LLMs for sentiment analysis is shown at Fig. 1.It describes every step of the procedure from quick construction and dataset preparation to LoRA-based fine-tuning and performance assessment.

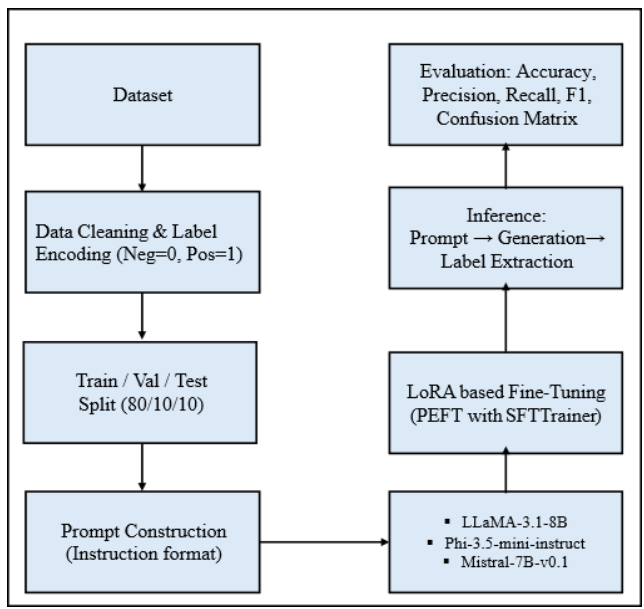

Fig. 1. Workflow for the study(LLMs)

Only the Review (input text) and Sentiment (target label) columns were used for fine-tuning. Each review was reformatted into a prompt-response format in order to modify the dataset for instruction-tuning. The input prompt for training and assessment prompted the model to categorize the sentiment of a particular review, and the right response was marked as supervision. The identical format was utilized for testing, but the gold label was omitted; thus, the model had to make the forecast. Example training prompt:

*Classify the following customer review into "Positive" or "Negative".*

*Return only the sentiment label.*

*[This product exceeded my expectations!] = Positive*

Example Testing prompt:

*Classify the following customer review into "Positive" or "Negative".*

*Return only the sentiment label.*

*[The quality was very poor.] =*

This prompt-based design aligns with recent instruction-tuning practices and allows the model to learn task-specific behavior in a generative setting. For model and quantization, we utilized Llama-3.1-8B, Phi-3.5-mini-instruct, Mistral-7B-v0.1; some state-of-the-art LLMs. To reduce memory consumption and enable efficient fine-tuning, we applied 4-bit quantization using the bitsandbytes library with the NF4 quantization scheme and float16 computation. This configuration allows large-scale models to be fine-tuned on commodity GPUs without significant performance degradation. For Parameter-Efficient Fine-Tuning (PEFT with LoRA), instead of full model fine-tuning, we adopted Low-Rank Adaptation (LoRA) under the PEFT framework to reduce computational overhead.LoRA introduces trainable rank-decomposition matrices into linear layers while freezing the base model parameters, enabling efficient adaptation.The LoRA configuration used rank (r) = 64,alpha = 16, target modules was all 4-bit linear layers except the final language modeling head. This setup significantly reduces the number of trainable parameters while maintaining high performance.

#### a) LLMs Selection

Three open-source LLMs have been selected for evaluation. The description of the LLMs are given below:

*Llama-3.1-8B :* Llama-3.1-8B is a foundation model for general-purpose natural language production and interpretation that was released by Meta in July 2024. It can handle lengthy documents efficiently since it has 8 billion parameters and an enlarged 128K-token context window. Distributed under the Llama 3.1 Community License, the model provides a flexible foundation for fine-tuning downstream in a variety of NLP tasks. Its robust zero-shot and few-shot performance is demonstrated by benchmarks, which makes it ideal for both practical and research applications in low-resource language environments.

*Phi-3.5-mini-instruct :* Microsoft created a lightweight, highly accurate instruct model that is a member of the Phi-3 family. It uses strict fine-tuning methods, such as supervised tuning, proximal policy optimization, and direct preference optimization, to guarantee robust command adherence and safety. It has 3.8 billion parameters and supports a 128K-token context. Its dataset, which aims for reasoning-quality instruction handling, includes both filtered and synthetic public online information.

*Mistral-7B-v0.1 :* Mistral AI produced a powerful, efficiency-optimized LLM with seven billion parameters. In reasoning, math, and code creation, it performs better than Llama 2 13B on a number of benchmarks. Its architecture makes use of sliding-window attention (SWA) and grouped-query attention (GQA) to enable quick and scalable inference. It is openly usable for a variety of applications and was released under Apache 2.0.

*2) BERT Models*

From dataset preparation and tokenization to model fine-tuning with several linguistic versions, the workflow presented at Fig. 2, demonstrates the experimental pipeline employing BERT-based models. Performance indicators like accuracy, precision, recall, F1 score, and confusion matrix are used in the evaluation phase. consecutively.

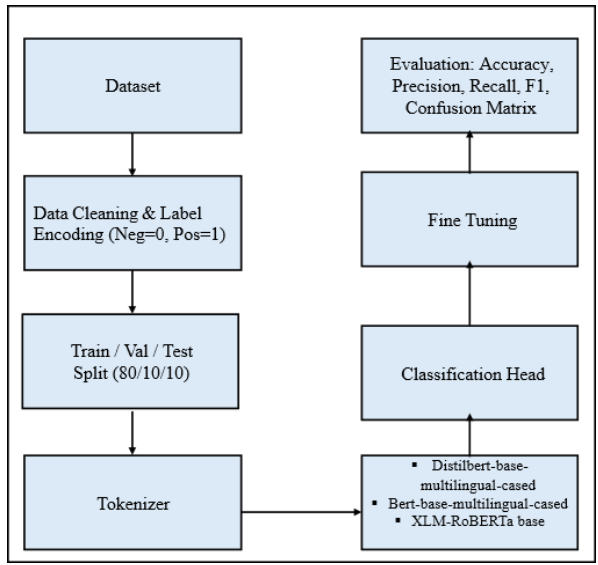

Fig. 2. Workflow for the study(BERT Models)

For BERT-based models, we adopted a transformer-based fine-tuning pipeline for sentiment classification of product reviews. Only the Review (input text) and Sentiment (target label) columns were used for fine-tuning. The methodology consists of some major steps. For text tokenization, we employed the chosen model's tokenizer. Each review was tokenized into input IDs and attention masks, truncated/padded to a maximum length of 128 tokens. We fine-tuned the models by attaching a classification head with two output units.

*a) BERT Models Selection*

Three open-source BERT models have been selected for evaluation. The description of the models are given below:

*DistilBERT-base-multilingual-cased(DistilBERT-multilingual) :* Hugging Face released a condensed, lightweight version of BERT that is intended for effective multilingual comprehension. It is appropriate for contexts with limited resources because it is 40% smaller and 60% faster while maintaining 97% of BERT's performance. Because the model is cased and supports 104 languages, it can be used for tasks like NER, classification, and sentiment analysis in a variety of linguistic contexts.

*BERT-base-multilingual-cased (mBERT) :* Google created a foundation transformer model that was pre-trained with 104-language Wikipedia dumps and a shared WordPiece vocabulary of 110k tokens. It features 12 attention heads (110M parameters), 12 layers, and 768 hidden units. For languages that are case-sensitive, the cased setting maintains the original text casing. For multilingual downstream tasks like categorization, question answering, and cross-lingual transfer, this paradigm is frequently utilized.

*XLM-RoBERTa base (XLM-R-base) :* This multilingual transformer, which was created by Facebook AI (Meta) as a scaled-down version of XLM-RoBERTa, has 12 layers and 270 million parameters, making it more effective in contexts with limited resources. It offers a balance between speed and accuracy, and like its larger version, it was trained on 2.5TB of Common Crawl material in 100 different languages. It performs better than mBERT on multilingual benchmarks like XNLI and MLQA, although being marginally less potent than XLM-R-large. It is frequently used for applications like sentiment analysis, classification, and cross-lingual transfer learning.

*b) Experimental Set Up*

Every experiment used an NVIDIA Tesla P100 GPU with 16 GB VRAM, 2 vCPUs, and 13 GB RAM, and was conducted on Kaggle's cloud environment. Hugging Face Transformers and PyTorch, two deep learning frameworks, were used in the Python 3.10 implementation. Pandas, NumPy, and Scikit-learn were used for data preprocessing and assessment, and Matplotlib was used for visualization. Random seeds were fixed for all libraries to guarantee reproducibility, and mixed precision training was used to maximize GPU memory utilization.

IV. RESULT AND DISCUSSION

Fine-tuning was conducted using the SFTTrainer from the Hugging Face TRL library for LLMs. Training hyperparameters are presented in TABLE II for LLMs. Validation was performed every 500 steps to monitor overfitting. All experiments were tracked with Weights & Biases for reproducibility. During inference, the model generated sentiment predictions from test prompts with a maximum of 5 new tokens and a temperature set to 0.1. Post-processing involved string matching to extract "Positive" or "Negative" labels. We evaluated model performance using overall accuracy, per-class accuracy, classification report (precision, recall, F1-score), and confusion matrix. This multi-metric evaluation provided a comprehensive understanding of classification robustness and error patterns.

TABLE II. HYPERPARAMETER SETTINGS FOR FINE TUNING LARGE LANGUAGE MODELS

| Hyperparameters | Values |
|---|---|
| Epochs | 1 |
| Per-device batch size | 1 |
| Gradient accumulation steps | 8 |
| Optimizer | paged_adamw_32bit |
| Learning rate | 2e-4 with cosine decay scheduler |
| Warmup ratio | 0.03 |
| Gradient checkpointing | enabled |
| Mixed precision | FP16 |

For BERT based models training was performed using Hugging Face's Trainer API with the following hyperparameters presented in TABLE III. To ensure reproducibility and disk safety, we disabled external logging (TensorBoard/W&B) and checkpoint saving.

TABLE III. HYPERPARAMETER SETTINGS FOR FINE TUNING BERT-BASED MODELS

| Hyperparameters | Values |
|---|---|
| Learning rate | 5e-5 |
| Batch size | 16 |
| Epochs | 3 |
| Weight decay | 0.01 |

After training and evaluation, the fine-tuned Llama outperformed all other models. The classwise performance is presented in TABLE IV for fine-tuned Llama-3.1-8B which performs comparably poorly on the Negative class (F1-score: 0.83), but remarkably well on the Positive class (F1-score: 0.97). Recall 0.88, F1-score 0.90, and accuracy 0.93 are all excellent average scores for the model.

TABLE IV. CLASSWISE PERFORMANCE OF FINE TUNED LLAMA-3.1-8B

| Class Name | Precision | Recall | F1-Score |
|---|---|---|---|
| Positive | 0.96 | 0.99 | 0.97 |
| Negative | 0.90 | 0.78 | 0.83 |
| **Average** | **0.93** | **0.88** | **0.90** |

The confusion matrix for sentiment classification with the fine tuned Llama-3.1-8B is shown in Fig. 3, which demonstrates that the model detected the majority of positive samples (337/342) correctly with very few misclassifications. Thirteen were incorrectly labeled as positive, but forty-five were accurately predicted as negative samples.

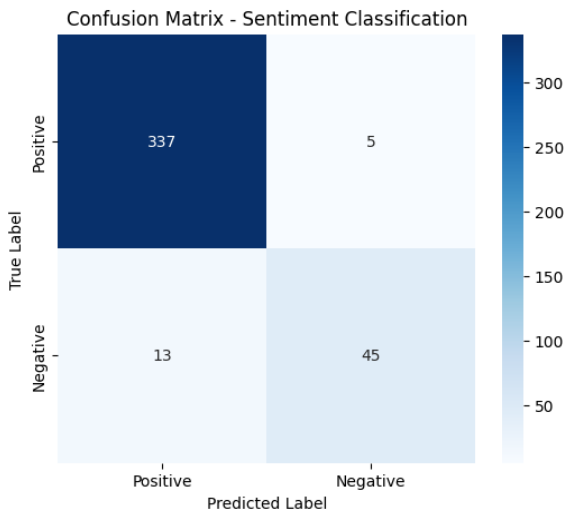

Fig. 3. Confusion matrix of fine-tuned Llama-3.1-8B

The overall sentiment classification performance of several models is compared in TABLE V. With an F1 score of 86%, XLM-R-base outperformed the other BERT-based models. The LLMs fared better than them; Llama-3.1 (8B) had the highest accuracy (95.5%) and F1 score (90%), indicating that it was better suited to handle sentiment analysis in Bangla.

A comparison with previous research, where Llama and mBERT models obtained accuracies below 90%, is shown in TABLE VI. They were greatly exceeded by the suggested improved LLaMA-3.1 (8B), which achieved 95.5% test accuracy, demonstrating its usefulness for sentiment analysis in Bangla. The experimental findings show that in Bangla-English mixed sentiment analysis, the fine-tuned Llama-3.1-8B performs noticeably better than BERT-based and other LLM baselines. Strong generalization across positive and negative classes is demonstrated by its exceptional precision and memory. The model's applicability for low-resource contexts is further demonstrated by the incorporation of parameter-efficient fine-tuning techniques like LoRA and PEFT. These results support LLMs' potential to further sentiment analysis studies in underrepresented languages, such as Bangla.

TABLE V. OVERALL PERFORMANCE COMPARISON BETWEEN ALL MODELS

| Model Name | Precision(%) | Recall(%) | F1 score(%) | Test Accuracy(%) |
|---|---|---|---|---|
| DistilBERT-multilingual | 86 | 80 | 82 | 92 |
| mBERT | 87 | 79 | 82 | 92 |
| XLM-R-base | 90 | 83 | 86 | 94 |
| Mistral-7B | 90 | 89 | 89 | 94.8 |
| Phi-3.5-mini | 90 | 77 | 81 | 92.2 |
| **Llama-3.1 8B** | **93** | **88** | **90** | **95.5** |

TABLE VI. RESULTS OF PREVIOUS STUDIES COMPARED WITH THE PROPOSED MODEL

| Reference | Model | Test Accuracy(%) |
|---|---|---|
| [5] | Llama | 89 |
| [7] | mBERT | 87 |
| [6] | Llama | 89.76 |
| Fine Tuned Llama-3.1 (8B) | Llama | 95.5 |

V. CONCLUSION

This study shows how well the LLaMA-3.1-8B model can be adjusted for sentiment analysis in the Bangla-English mixed reviews, which have limited resources. The fine-tuned model Llama-3.1 (8B) outperformed BERT-based and other LLM baselines with an overall accuracy of 95.5% and good precision, recall, and F1-score of about 93%, 88%, 90%. The incorporation of parameter-efficient fine-tuning techniques like PEFT and LoRA demonstrates the model's versatility in settings with limited resources. These findings demonstrate that LLMs can be effectively modified for specific NLP applications in low-resource languages such as Bangla. However, more research using bigger datasets and more diverse sentiment categories may improve results, especially for minority classes. To enhance generalization, future research can concentrate on gathering larger datasets from other Bangladeshi e-commerce platforms. Higher accuracy can also be achieved by fine-tuning LLMs on individual language data and experimenting with more complex language classifiers. However, this study was restricted to a subset of data due to computational and resource limitations, allowing for additional investigation with more extensive resources.